\documentclass[preprint,12pt]{elsarticle}
\usepackage[a4paper,margin=25mm]{geometry}
\usepackage{amsmath,amssymb,amsthm}
\usepackage{booktabs,multirow,tabularx,array}
\usepackage{algorithm,algpseudocode}
\usepackage{graphicx,xcolor,tikz}
\usetikzlibrary{arrows.meta,positioning,fit,backgrounds,calc}
\usepackage{microtype}
\usepackage[section]{placeins}
\usepackage[hidelinks]{hyperref}
\hypersetup{pdftitle={TRCM: Topology-aware route-consistency modeling for autoregressive neural vehicle routing},pdfauthor={Peiheng Jian, Zhenwei Wang, Tiehua Zhang, Lei Zhang, Zhishu Shen, Ruibin Bai, Zhijun Ding},pdfsubject={Neurocomputing manuscript}}
\graphicspath{{figures/}}
\journal{Neurocomputing}
\newcommand{\trcm}{\textsc{TRCM}}
\newcommand{\Vc}{\mathcal V_c}

\newcommand{\ind}{\mathbb I}
\newcommand{\BCE}{\operatorname{BCE}}
\newcommand{\TE}{\operatorname{TE}}
\newcommand{\softmax}{\operatorname{softmax}}
\newtheorem{proposition}{Proposition}
\begin{document}
\begin{frontmatter}
\title{TRCM: Topology-aware route-consistency modeling for autoregressive neural vehicle routing}
\author[tongji]{Peiheng~Jian\fnref{equal}}
\author[nottingham]{Zhenwei~Wang\fnref{equal}}
\author[tongji]{Tiehua~Zhang\corref{corresponding}}
\ead{tiehuaz@tongji.edu.cn}
\author[yixin]{Lei~Zhang}
\author[wuhan]{Zhishu~Shen}
\author[nottingham]{Ruibin~Bai}
\author[tongji]{Zhijun~Ding}
\affiliation[tongji]{organization={School of Computer Science and Technology, Tongji University},city={Shanghai},postcode={201804},country={China}}
\affiliation[nottingham]{organization={School of Computer Science, University of Nottingham Ningbo China},city={Ningbo},postcode={315100},country={China}}
\affiliation[yixin]{organization={Yixin Group Limited},city={Beijing},country={China}}
\affiliation[wuhan]{organization={School of Computer Science and Artificial Intelligence, Wuhan University of Technology},city={Wuhan},postcode={430070},country={China}}
\fntext[equal]{These authors contributed equally to this work.}
\cortext[corresponding]{Corresponding author.}

\begin{abstract}
Autoregressive neural solvers construct capacitated vehicle routing solutions through sequential node selection, while the partition of customers into vehicle routes remains implicit. We propose Topology-aware Route-Consistency Modeling (TRCM) to represent customer co-assignment explicitly and use it to guide construction. A symmetric pair predictor reuses encoder representations to estimate same-route probabilities. Reference-preserving augmentation and topological entropy-based reweighting address imbalanced relation supervision. During decoding, compatibility with the current partial route adjusts customer logits while the base solver retains its feasibility and state-update rules. Under matched decoding settings, TRCM reduces the reported gap from 3.67\% to 3.28\% on CVRP50 and from 2.11\% to 1.99\% on CVRP100. With eightfold augmentation, the CVRP100 gap decreases from 1.61\% to 1.49\%, although performance deteriorates on CVRP50. Across 10,000 XML100 instances, the average objective improves by 0.24\%, with larger differences under unitary demands, clustered customers and ultra-long routes. A controlled study of 500 demand-reassigned variants of one geometry also shows lower objective variation. The results support route membership as a useful, condition-dependent structural signal for autoregressive neural routing.
\end{abstract}
\begin{keyword}
Capacitated vehicle routing \sep Neural combinatorial optimization \sep Route consistency \sep Pairwise learning \sep Topological entropy \sep Autoregressive decoding
\end{keyword}
\end{frontmatter}

\section{Introduction}
\label{sec:introduction}

The capacitated vehicle routing problem (CVRP) is a central problem in logistics and transportation: customers with known demands must be assigned to capacity-feasible vehicle routes of minimum total length~\cite{toth2014vehicle}. Classical heuristics and metaheuristics offer high-quality solutions through specialized search~\cite{lkh3,hgs,Wouda2023ALNS,Wouda2024PyVRP}. Neural combinatorial optimization (NCO) complements these methods by learning reusable construction policies from problem instances and solutions~\cite{bello2017nco,nazari2018vrp,kool2019attention}. Once trained, a policy can generate feasible solutions through a sequence of inexpensive decisions.

Autoregressive construction is a common interface for neural routing. An encoder represents the instance, and a decoder selects the next customer or the depot using the partial solution and a feasibility mask. Attention-based models and multiple-start training have made this interface effective on standard CVRP benchmarks~\cite{kool2019attention,kwon2020pomo}. Subsequent work improves symmetry exploitation, decoder capacity, reconstruction, and inference~\cite{Kim2022SymNCO,luo2023lehd,huang2025reld,choo2022sgbs}. These developments strengthen the decisions made during construction while retaining a sequential action interface.

\begin{figure}[p]
\centering
\includegraphics[width=\linewidth]{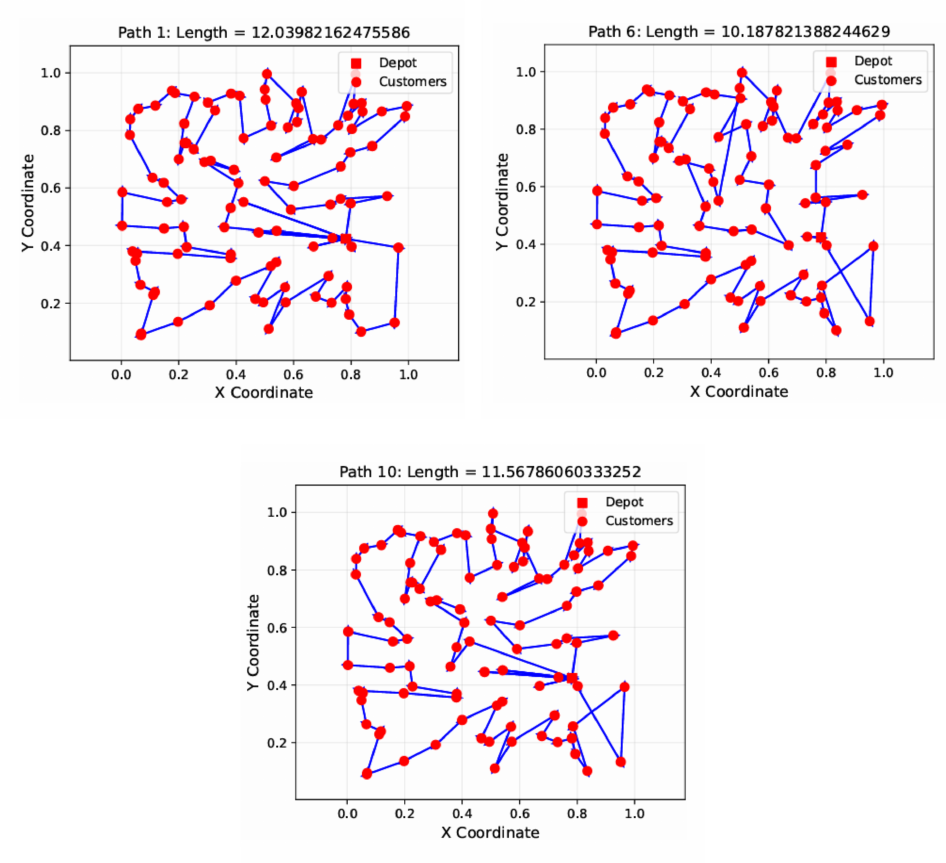}
\caption{Constructive decoding trajectories for three selected demand-reassigned variants at fixed customer coordinates. Visiting trajectories and route lengths vary across the examples.}
\label{fig:motivation_case}
\end{figure}

A CVRP solution, however, contains a structural object beyond its visiting sequence: the partition of customers among vehicle routes. In a node-level policy, this partition is represented implicitly through embeddings, decoder states, and transition preferences. An explicit representation of customer co-assignment could provide a complementary signal about the route being constructed. Figure~\ref{fig:motivation_case} shows three selected demand-reassigned variants at fixed customer coordinates. The examples illustrate how a fixed spatial configuration can yield different constructive trajectories under demand reassignment, motivating explicit modeling of route-level organization.

We consider whether two customers belong to the same route. This same-route relation differs from edge adjacency: co-assigned customers need not be visited consecutively. It is also invariant to reordering visits within a reference route, making it a natural representation of route membership. Learning it as a soft pairwise prior allows the constructive policy to query route compatibility while continuing to choose individual actions.

Three challenges arise. First, same-route pairs are usually a minority among all customer pairs, creating imbalanced supervision. Second, pairs with the same label can have different relation-label contexts, so class counts alone do not describe their learning difficulty. Third, a route-level signal must be integrated with a decoder that already models transition costs and dynamic constraints. Edge-classification research motivates considering local label context in addition to frequency~\cite{cheng2025topoedge}; the integration requires a separate connection between pair predictions and the current partial route.

We propose Topology-aware Route-Consistency Modeling (\trcm). A symmetric pair predictor reuses encoder representations to estimate customer same-route probabilities. Prior learning combines reference-preserving augmentation with topological entropy (TE) and class-frequency weighting. During decoding, \trcm{} averages each candidate's predicted compatibility with customers already in the current route and adds this score to its base logit. The original solver maintains the dynamic state, feasibility mask, and action-selection procedure. This connects an order-invariant relation target to sequential routing without requiring a separately decoded partition.

We evaluate the resulting POMO-based solver on CVRP50, CVRP100, and all 10,000 XML100 instances. Under matched decoding settings, \trcm{} reduces the non-augmented gaps from 3.67\% to 3.28\% and from 2.11\% to 1.99\%, respectively. With eightfold augmentation, it improves CVRP100 and degrades CVRP50, indicating that the interaction with stronger inference is not uniformly beneficial. XML100 reveals larger gains for unitary demands, clustered customers, and ultra-long routes. A controlled study of 500 demand-reassigned variants of one geometry shows lower objective variation. Prior-prediction ablations, sensitivity analysis, and measured overhead further characterize the method.

\begin{samepage}
Our contributions are threefold:
\begin{enumerate}
\item We formulate route membership as an explicit pairwise structural representation for autoregressive CVRP construction, distinguishing customer co-assignment from adjacency and visiting order.
\item We develop topology-aware route-consistency learning and route-context-guided decoding through symmetric pair prediction, imbalance-aware supervision, and additive logit fusion.
\item We evaluate the method using standard benchmarks, structured XML100 factors, controlled demand reassignment, prior-learning ablations, hyperparameter sensitivity, and computational overhead.
\end{enumerate}
\end{samepage}

\section{Related Work}
\label{sec:related}

\subsection{Neural constructive solvers for vehicle routing}

Neural routing policies combine representations of the instance with a learned construction rule. Pointer Networks introduced attention-based selection over input elements~\cite{vinyals2015pointer}; reinforcement learning extended this principle to combinatorial optimization and demand-dependent routing states~\cite{bello2017nco,nazari2018vrp}. Transformer-style attention and multiple-start optimization subsequently became common components of constructive solvers~\cite{kool2019attention,kwon2020pomo}. Improvements to this family address different bottlenecks: symmetry-based training regularizes equivalent solutions~\cite{Kim2022SymNCO}, heavy decoders update representations of remaining nodes~\cite{luo2023lehd}, and revised light decoders strengthen the use of context~\cite{huang2025reld}. Large-scale training and reconstruction provide another direction~\cite{luo2025boosting}. Most of these approaches organize construction through successive next-node decisions. Insertion-based construction instead selects both a customer and an insertion position~\cite{luo2025insert}, illustrating that the construction interface itself can also be redesigned.

Generalization methods address the distribution or task on which these policies operate. Policy ensembles combine complementary behaviors across distributions~\cite{jiang2023ensemble,gao2023elg}. Cross-problem learning transfers a shared backbone to routing variants~\cite{lin2024crossproblem}, while prompt learning adapts a pretrained routing model through selected prompts~\cite{liu2024prompt}. Recent Neurocomputing work also studies route diversity and complementary distance-aware branches~\cite{guan2025diverse,jiang2026dbda}. \trcm{} focuses on the structural target used within construction: customer co-assignment is learned explicitly and queried through the current route.

\subsection{Structural and relational modeling in neural routing}

Relational information enters neural routing at several levels. Residual edge-graph attention and neighborhood-based edge fusion enrich node representations with graph and edge features~\cite{lei2022egat,wang2024gase}. Edge-based generation, including graph diffusion, models a different representation of the solution~\cite{sun2023difusco}. These approaches demonstrate the usefulness of pairwise structure, but edge adjacency and route membership are distinct targets. A same-route label also covers customers separated by several visits. Existing representations may encode route information implicitly; they do not explicitly optimize the same-route co-assignment target used here.

Learned structural guidance can also act within an optimization procedure. Neural large-neighborhood search learns repair operations~\cite{hottung2020nlns}, delegation selects subproblems for improvement~\cite{Li2021Delegate}, and NeuroLKH uses learned scores to guide classical search~\cite{Xin2021NeuroLKH}. Simulation-guided beam search and efficient active search increase inference effort through rollout or adaptation~\cite{choo2022sgbs,hottung2022eas}. \trcm{} supplies its relation signal directly to autoregressive action scoring, adding pairwise computation without introducing a separate search stage.

\subsection{Imbalance-aware pairwise learning}

Imbalanced supervision arises when the relations of interest occupy a small fraction of possible pairs. Focal Loss changes the emphasis on well-classified examples~\cite{lin2017focalloss}, while Class-Balanced Loss uses effective class counts~\cite{cui2019classbalanced}. Edge-representation learning treats relations as explicit prediction objects~\cite{bielak2022attre2vec}. TopoEdge further studies imbalance in local edge-label context and introduces TE-based weighting~\cite{cheng2025topoedge}. These perspectives distinguish sample frequency, prediction difficulty, and graph context.

For route-consistency learning, \trcm{} combines inverse-frequency weights with TE computed from supervised same-route labels. TE measures the mixture of relation labels around a pair's endpoints after propagation. Its purpose is to adjust learning emphasis for the relation predictor; it is not an estimator of route optimality.

\section{Problem Formulation}
\label{sec:formulation}

\subsection{Capacitated vehicle routing problem}

Let $\mathcal G=(\mathcal V,\mathcal E)$ be a complete routing graph with depot $0$, customer set $\Vc=\{1,\ldots,n\}$, customer demands $d_i$, vehicle capacity $Q$, and travel costs $c_{ij}$. A route $R_k=(0,v_{k,1},\ldots,v_{k,\ell_k},0)$ serves the customer set $\mathcal C(R_k)$. A feasible solution $\mathcal R$ partitions the customers into capacity-feasible routes:
\begin{equation}
\bigcup_{R\in\mathcal R}\mathcal C(R)=\Vc,\quad
\mathcal C(R_k)\cap\mathcal C(R_l)=\varnothing\ (k\ne l),\quad
\sum_{i\in\mathcal C(R)}d_i\le Q.
\label{eq:feasibility}
\end{equation}
The objective is the total route length,
\begin{equation}
\min_{\mathcal R}\sum_{R_k\in\mathcal R}\left(c_{0,v_{k,1}}+\sum_{a=1}^{\ell_k-1}c_{v_{k,a},v_{k,a+1}}+c_{v_{k,\ell_k},0}\right).
\label{eq:cvrp_objective}
\end{equation}
The evaluated synthetic instances use Euclidean travel costs.

\subsection{Autoregressive neural construction}

An encoder produces node representations $\mathbf H=\operatorname{Enc}_\theta(\mathcal G)$. A decoder constructs a trajectory $\tau$ according to
\begin{equation}
p_\theta(\tau\mid\mathcal G)=\prod_{t=1}^{T}p_\theta(\tau_t\mid\mathcal G,\tau_{<t}).
\label{eq:prob_chain}
\end{equation}
At step $t$, its base logits $u_t(j)$ depend on the embeddings and current solver state. The feasible set $\mathcal F_t$ excludes served and capacity-infeasible customers. Selecting the depot closes the current route and restores capacity according to the base solver. Multi-start decoding and test-time augmentation can produce several trajectories for the same instance.

\subsection{Route membership versus edge adjacency}

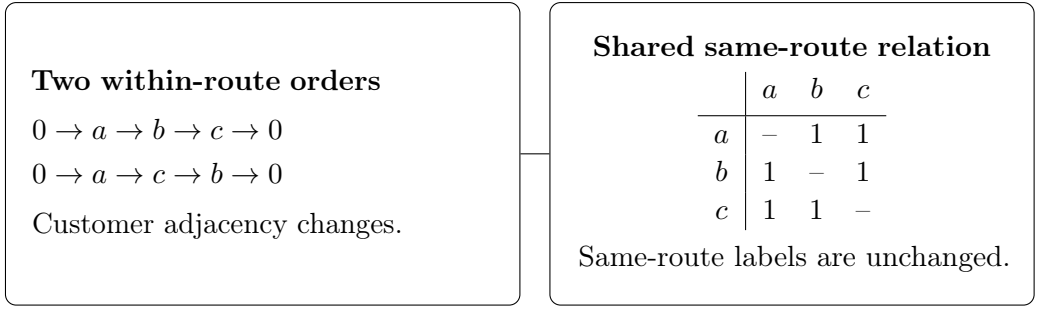
\begin{figure}[t]
\centering
\begin{tikzpicture}[font=\small,>=Latex]
\node[draw,rounded corners,align=left,text width=6.1cm,minimum height=4cm,inner sep=10pt] (a) at (0,0) {\textbf{Two within-route orders}\\[5pt]
$0\to a\to b\to c\to0$\\[3pt]
$0\to a\to c\to b\to0$\\[5pt]
Customer adjacency changes.};
\node[draw,rounded corners,align=center,text width=5.7cm,minimum height=4cm,inner sep=10pt] (b) at (7.0,0) {\textbf{Shared same-route relation}\\[5pt]
\begingroup\setlength{\arraycolsep}{6pt}\renewcommand{\arraystretch}{1.05}%
$\begin{array}{c|ccc}&a&b&c\\[2pt]\hline a&\text{\textendash}&1&1\\b&1&\text{\textendash}&1\\c&1&1&\text{\textendash}\end{array}$\endgroup\\[5pt]
Same-route labels are unchanged.};
\draw[line width=0.4pt] (a.east) -- (b.west);
\end{tikzpicture}
\caption{Visiting order and route membership are different structural objects. Reordering customers $a,b,c$ changes their adjacency relations while preserving their same-route labels.}
\label{fig:structural_levels}
\end{figure}

A reference solution $\mathcal R^\star$ defines customer-pair labels
\begin{equation}
y_{ij}=\ind\{\exists R\in\mathcal R^\star:i,j\in\mathcal C(R)\},\qquad i,j\in\Vc.
\label{eq:same_route_label}
\end{equation}
The label depends on route membership, not on whether $i$ and $j$ are consecutive. Permuting visits within a reference route leaves every $y_{ij}$ unchanged, as illustrated in Figure~\ref{fig:structural_levels}. \ref{app:analysis} gives this property together with the symmetry and aggregation results used below.

The predictor estimates a customer-only matrix $\mathbf{P}=[P_{ij}]\in[0,1]^{n\times n}$, where $P_{ij}$ represents the probability of same-route co-assignment. The depot is excluded because all vehicle routes visit it. A real-valued predicted matrix need not be exactly transitive or define a feasible partition. We therefore use $\mathbf{P}$ as a soft route-consistency prior queried during construction.

\begin{table}[htbp]
\centering\small
\caption{Main notation. The encoder parameters $\theta$ and the reweighting balance $\beta$ are distinct.}
\label{tab:notation}
\begin{tabularx}{\linewidth}{lX}
\toprule
Symbol & Meaning \\
\midrule
$\Vc,0$ & Customer set and depot \\
$\mathcal{R}^{\star},y_{ij}$ & Reference route solution and same-route label \\
$\mathcal{P}$ & Supervised customer-pair set for training \\
$\mathbf{P}=[P_{ij}]$ & Dense route-consistency prior matrix \\
$\mathbf{H},g_\phi$ & Encoder representations and symmetric pair predictor \\
$\widetilde{\mathbf{A}}$ & Row-normalized relation-graph adjacency with self-loops \\
$\alpha,L,\beta$ & TE temperature, propagation depth, and weighting balance \\
$\mathcal{S}_t,\mathcal{F}_t$ & Customers in the current partial route and feasible actions \\
$u_t(j),r_t(j),\lambda$ & Base logit, route-compatibility score, and fusion strength \\
\bottomrule
\end{tabularx}
\end{table}

\section{Topology-Aware Route-Consistency Modeling}
\label{sec:method}

\subsection{Framework overview}

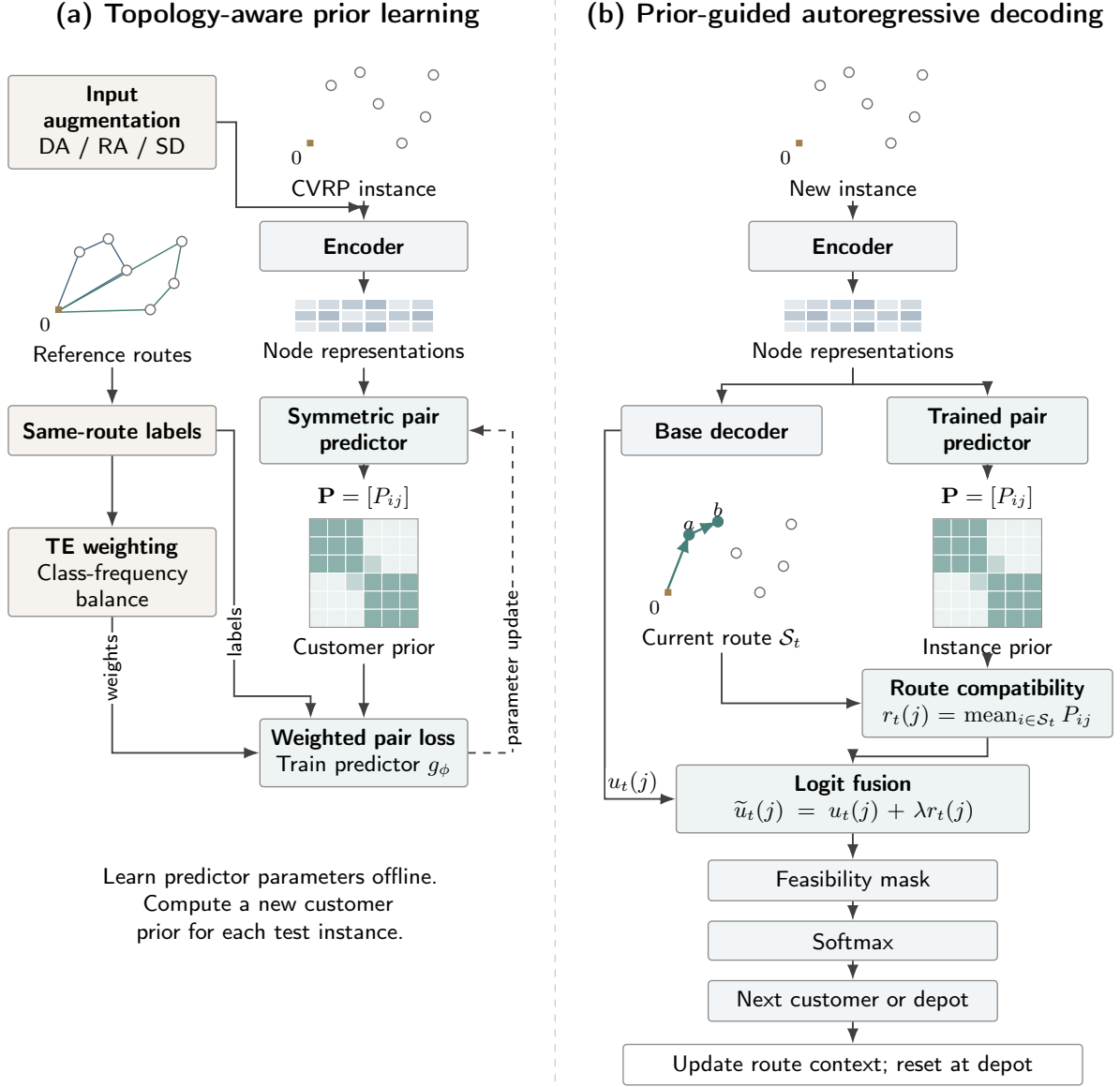
\begin{figure}[p]
\centering
\begingroup
\definecolor{trcmBlue}{RGB}{76,108,136}
\definecolor{trcmTeal}{RGB}{69,127,119}
\definecolor{trcmOchre}{RGB}{161,128,76}
\newcommand{\trcmGraph}[1]{%
\begin{tikzpicture}[x=.37cm,y=.37cm,>=Latex,line width=.55pt]
\coordinate (z) at (0,0);\coordinate (a) at (.9,2.3);
\coordinate (b) at (2,2.8);\coordinate (c) at (2.7,1.6);
\coordinate (d) at (3.6,.1);\coordinate (e) at (4.5,1.1);\coordinate (f) at (4.8,2.7);
\ifnum#1=1
\draw[trcmBlue] (z)--(a)--(b)--(c)--(z);
\draw[trcmTeal] (z)--(d)--(e)--(f)--(z);
\fi
\ifnum#1=2
\draw[->,trcmTeal,line width=.9pt] (z)--(a);
\draw[->,trcmTeal,line width=.9pt] (a)--(b);
\fi
\foreach \p in {a,b,c,d,e,f}{\filldraw[fill=white,draw=black!55] (\p) circle (2pt);}
\ifnum#1=2
\foreach \p in {a,b}{\filldraw[fill=trcmTeal,draw=trcmTeal] (\p) circle (2pt);}
\node[font=\fontsize{8}{9}\selectfont,anchor=south] at (a) {$a$};
\node[font=\fontsize{8}{9}\selectfont,anchor=south] at (b) {$b$};
\fi
\filldraw[fill=trcmOchre,draw=trcmOchre] (z) rectangle ++(.2,.2);
\node[font=\fontsize{8}{9}\selectfont,anchor=north east] at (z) {$0$};
\path[use as bounding box] (-.45,-.35) rectangle (5.1,3.15);
\end{tikzpicture}}
\newcommand{\trcmMatrix}{%
\begin{tikzpicture}[x=.255cm,y=.255cm]
\foreach \i in {0,...,5}{\foreach \j in {0,...,5}{%
\pgfmathtruncatemacro{\same}{int(\i/3)==int(\j/3)}
\ifnum\same=1\def\shade{62}\else\def\shade{10}\fi
\pgfmathtruncatemacro{\ambiguous}{(\i==2 && \j==3)||(\i==3 && \j==2)}
\ifnum\ambiguous=1\def\shade{32}\fi
\fill[trcmTeal!\shade] (\i,-\j) rectangle ++(1,-1);
\draw[white,line width=.35pt] (\i,-\j) rectangle ++(1,-1);
}}
\draw[black!45,line width=.4pt] (0,0) rectangle (6,-6);
\end{tikzpicture}}
\newcommand{\trcmEmbedding}{%
\begin{tikzpicture}[x=.33cm,y=.16cm]
\foreach \i in {0,...,5}{\foreach \j in {0,...,2}{%
\pgfmathtruncatemacro{\shade}{15+mod(\i+2*\j,4)*10}
\fill[trcmBlue!\shade] (\i,-\j) rectangle ++(.87,-.8);
}}
\end{tikzpicture}}
\begin{tikzpicture}[
font=\fontsize{9}{10.5}\selectfont\sffamily,>=Latex,
every node/.style={execute at begin node={\hyphenpenalty=10000\exhyphenpenalty=10000}},
box/.style={draw=black!45,rounded corners=1.5pt,line width=.5pt,align=center,inner sep=4pt,text width=2.65cm,minimum height=.7cm},
base/.style={box,fill=trcmBlue!7},prior/.style={box,fill=trcmTeal!9},
sup/.style={box,fill=trcmOchre!10},
object/.style={align=center,inner sep=2pt},arr/.style={->,line width=.7pt,draw=black!75}]
\path[use as bounding box] (0,-15.2) rectangle (15.95,.5);
\draw[black!20,dashed,line width=.5pt] (7.9,.4)--(7.9,-15.05);
\node[align=center,text width=7.5cm,font=\fontsize{10.5}{12}\selectfont\sffamily\bfseries] at (3.85,.15) {(a) Topology-aware prior learning};
\node[align=center,text width=7.6cm,font=\fontsize{10.5}{12}\selectfont\sffamily\bfseries] at (12.0,.15) {(b) Prior-guided autoregressive decoding};

\node[object] (instance) at (5.2,-1.45) {\trcmGraph{0}\\CVRP instance};
\node[sup] (aug) at (1.65,-1.35) {\textbf{Input augmentation}\\DA / RA / SD};
\node[base] (encL) at (5.2,-3.1) {\textbf{Encoder}};
\node[object] (embL) at (5.2,-4.3) {\trcmEmbedding\\Node representations};
\node[prior] (headL) at (5.2,-5.7) {\textbf{Symmetric pair}\\\textbf{predictor}};
\node[object] (matrixL) at (5.2,-7.7) {$\mathbf{P}=[P_{ij}]$\\[3pt]\trcmMatrix\\Customer prior};
\node[prior] (loss) at (5.2,-10.25) {\textbf{Weighted pair loss}\\Train predictor $g_\phi$};
\node[object] (reference) at (1.65,-3.8) {\trcmGraph{1}\\Reference routes};
\node[sup] (labels) at (1.65,-5.7) {\textbf{Same-route labels}};
\node[sup] (weights) at (1.65,-7.7) {\textbf{TE weighting}\\Class-frequency balance};
\draw[arr] (instance)--(encL);\draw[arr] (encL)--(embL);
\draw[arr] (embL)--(headL);\draw[arr] (headL)--(matrixL);\draw[arr] (matrixL)--(loss);
\draw[arr] (aug.east)--(3.4,-1.35)--(3.4,-2.55)--(5.2,-2.55);
\fill[black!75] (5.2,-2.55) circle (1pt);
\draw[arr] (reference)--(labels);\draw[arr] (labels)--(weights);
\draw[arr] (weights.south)--(1.65,-10.25)--(loss.west);
\node[fill=white,inner sep=1pt,font=\fontsize{8}{9}\selectfont\sffamily,rotate=90] at (1.65,-9.05) {weights};
\draw[arr] (labels.east)--(3.37,-5.7)--(3.37,-9.45)--(4.45,-9.45)--($(loss.north)+(-.75,0)$);
\node[fill=white,inner sep=1pt,font=\fontsize{8}{9}\selectfont\sffamily,rotate=90] at (3.37,-8.55) {labels};
\draw[arr,dashed] (loss.east)--(7.3,-10.25)--(7.3,-5.7)--(headL.east);
\node[fill=white,inner sep=1pt,font=\fontsize{8}{9}\selectfont\sffamily,rotate=90] at (7.3,-9.05) {parameter update};
\node[align=center,text width=6.7cm,font=\fontsize{9}{11}\selectfont\sffamily] at (3.85,-12.4) {Learn predictor parameters offline.\\Compute a new customer prior for each test instance.};

\node[object] (new) at (12.1,-1.45) {\trcmGraph{0}\\New instance};
\node[base] (encR) at (12.1,-3.1) {\textbf{Encoder}};
\node[object] (embR) at (12.1,-4.3) {\trcmEmbedding\\Node representations};
\node[base,text width=2.55cm] (decoder) at (10.25,-5.7) {\textbf{Base decoder}};
\node[prior,text width=2.55cm] (headR) at (14.0,-5.7) {\textbf{Trained pair predictor}};
\node[object] (matrixR) at (14.0,-7.7) {$\mathbf{P}=[P_{ij}]$\\[3pt]\trcmMatrix\\Instance prior};
\node[object] (route) at (10.25,-7.7) {\trcmGraph{2}\\Current route $\mathcal S_t$};
\node[prior,text width=3.25cm] (score) at (14.0,-9.55) {\textbf{Route compatibility}\\$r_t(j)=\operatorname{mean}_{i\in\mathcal S_t}P_{ij}$};
\node[prior,text width=4.7cm] (fusion) at (12.1,-10.9) {\textbf{Logit fusion}\\$\widetilde u_t(j)=u_t(j)+\lambda r_t(j)$};
\node[base,text width=3.8cm,minimum height=.55cm] (mask) at (12.1,-12.05) {Feasibility mask};
\node[base,text width=3.8cm,minimum height=.55cm] (softmax) at (12.1,-12.9) {Softmax};
\node[base,text width=3.8cm,minimum height=.55cm] (next) at (12.1,-13.75) {Next customer or depot};
\node[box,text width=6.2cm,minimum height=.55cm] (update) at (12.1,-14.65) {Update route context; reset at depot};
\draw[arr] (new)--(encR);\draw[arr] (encR)--(embR);
\draw[arr] (embR.south)--(12.1,-5.05)--(10.25,-5.05)--(decoder.north);
\draw[arr] (embR.south)--(12.1,-5.05)--(14.0,-5.05)--(headR.north);
\draw[arr] (headR)--(matrixR);\draw[arr] (matrixR)--(score);
\draw[arr] (route.south)--(10.25,-9.55)--(score.west);
\draw[arr] (decoder.west)--(8.6,-5.7)--(8.6,-10.9)--(fusion.west);
\node[anchor=south west,fill=white,inner sep=1pt] at (8.6,-10.87) {$u_t(j)$};
\draw[arr] (score.south)--(14.0,-10.3)--(12.1,-10.3)--(fusion.north);
\draw[arr] (fusion)--(mask);\draw[arr] (mask)--(softmax);
\draw[arr] (softmax)--(next);\draw[arr] (next)--(update);
\end{tikzpicture}
\endgroup
\caption{TRCM architecture. (a) Augmented inputs and reference-route labels train a symmetric pair predictor with topology and class-frequency weighting. (b) Shared node representations support base decoding and instance-specific prior prediction. The prior $\mathbf{P}$ and current route $\mathcal S_t$ determine compatibility scores for logit fusion, followed by feasibility masking and action selection. A depot return resets the route context. Graphs and matrix shading are schematic.}
\label{fig:framework}
\end{figure}

\trcm{} connects offline relation learning with autoregressive construction (Figure~\ref{fig:framework}). Reference routes provide same-route labels for an auxiliary pair predictor. Augmentation enriches its inputs, and topology and frequency weights modulate the pairwise learning objective. At inference, the trained predictor computes a new matrix for the current instance. Compatibility with the current route then complements the base decoder's logits.

\FloatBarrier

\subsection{Same-route relation prediction}

For customer embeddings $\mathbf h_i$ and $\mathbf h_j$, the predictor uses
\begin{equation}
\mathbf z_{ij}=[\mathbf h_i+\mathbf h_j\,\|\,|\mathbf h_i-\mathbf h_j|\,\|\,\mathbf h_i\odot\mathbf h_j],
\qquad P_{ij}=g_\phi(\mathbf z_{ij}),
\label{eq:pair_feature}
\end{equation}
where $\|$ is concatenation and $\odot$ is element-wise multiplication. The head $g_\phi$ is a two-layer MLP followed by a sigmoid. Each feature operation is symmetric, giving $P_{ij}=P_{ji}$. This symmetry concerns exchanging pair endpoints; the reference label's invariance concerns reordering an entire route. Both support a relation representation that is independent of a chosen serialization.

\subsection{Structure-preserving augmentation}

Training instances are paired with high-quality reference routes. The augmentation set contains geometric augmentation (DA), route-refactor augmentation (RA), and shuffle-demand augmentation (SD). DA includes rotation, reflection, shrinking or scaling, and small coordinate perturbations. RA uses reference-route refactoring inspired by partial-solution reconstruction~\cite{luo2023lehd}. SD permutes customer demands within each reference route while retaining its co-assignment labels.

The preserved property depends on the transformation. Rotation and reflection preserve Euclidean distances, and uniform scaling preserves route-length rankings. Within-route demand permutation preserves each reference route's total load and therefore its capacity feasibility. Small coordinate perturbations retain reference membership labels as training targets. Augmentation thus supplies varied inputs for learning the reference relation without requiring a unique visiting order.

\subsection{Topological entropy reweighting}
\label{sec:te}

Let $\mathcal{P}$ denote the supervised pair set, with $\mathcal{P}_c=\{(i,j)\in\mathcal{P}:y_{ij}=c\}$. Same-route pairs can be rare: for reference route sizes $\ell_1,\ldots,\ell_m$, their fraction among distinct unordered pairs is $\sum_k\ell_k(\ell_k-1)/[n(n-1)]$. Frequency weights address this imbalance. TE additionally describes the relation-label context of each pair~\cite{cheng2025topoedge}.

\begin{figure}[t]
\centering
\includegraphics[width=0.77\linewidth]{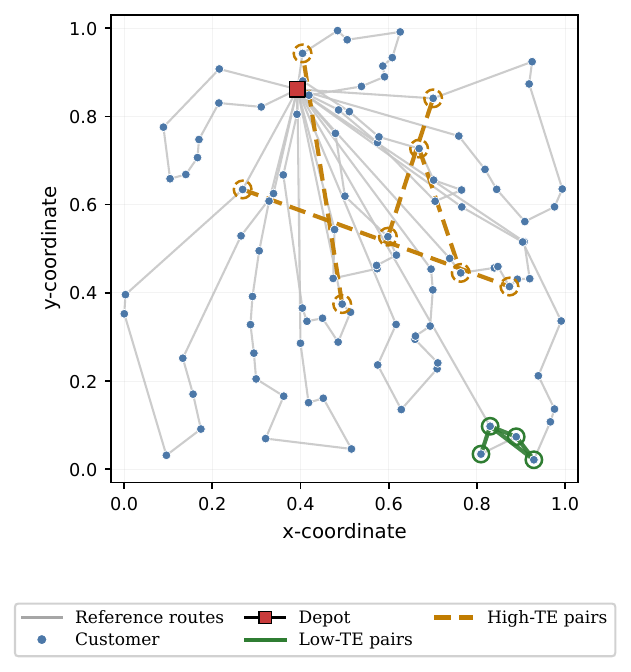}
\caption{Qualitative pair contexts for TE-based supervision. Green solid and orange dashed links indicate low- and high-TE examples, respectively. Gray lines show reference routes, blue points denote customers, and the red square is the depot.}
\label{fig:te_case}
\end{figure}

Construct the customer relation graph from supervised pairs, treating the endpoint relation symmetrically. For a node incident to supervision, define
\begin{equation}
q_{i,c}^{(0)}=\frac{\sum_{j:(i,j)\in\mathcal{P}}\ind\{y_{ij}=c\}}{\sum_{j:(i,j)\in\mathcal{P}}1},\qquad c\in\{0,1\}.
\label{eq:local_relation_distribution}
\end{equation}
Collect these distributions in $\mathbf Q^{(0)}$. With row-normalized relation adjacency $\widetilde{\mathbf A}$ including self-loops, propagate
\begin{equation}
\mathbf Q^{(l)}=\widetilde{\mathbf A}\mathbf Q^{(l-1)},\qquad l=1,\ldots,L.
\label{eq:te_propagation}
\end{equation}
The pair context and entropy are
\begin{align}
\mathbf g_{ij}^{(L)}&=\tfrac12(\mathbf q_i^{(L)}+\mathbf q_j^{(L)}),\label{eq:edge_context_dist}\\
\TE_{ij}^{(L)}&=-\sum_{c\in\{0,1\}}g_{ij,c}^{(L)}\log(g_{ij,c}^{(L)}+\epsilon),\label{eq:te}
\end{align}
where $\epsilon$ stabilizes the logarithm. Higher entropy indicates a more mixed endpoint relation-label context (Figure~\ref{fig:te_case}), rather than geometric boundary distance or route optimality. The supervised relation graph defines this context separately from the dense prediction matrix $\mathbf{P}=[P_{ij}]$.

The unnormalized weights are
\begin{equation}
w_{ij}^{\mathrm{TE}}=\exp(\TE_{ij}^{(L)}/\alpha),\qquad
w_{ij}^{\mathrm{qty}}=\frac{|\mathcal{P}|}{|\mathcal{P}_{y_{ij}}|},
\label{eq:raw_weights}
\end{equation}
where $\alpha>0$ controls the topology-weight contrast. Normalize each type by its mean over supervised pairs in the mini-batch:
\begin{equation}
\bar w_{ij}^{a}=\frac{w_{ij}^{a}}{|\mathcal{P}|^{-1}\sum_{(p,q)\in\mathcal{P}}w_{pq}^{a}},\qquad a\in\{\mathrm{TE},\mathrm{qty}\}.
\label{eq:weight_normalization}
\end{equation}
The final weighting and prior-learning loss are
\begin{align}
w_{ij}&=\beta\bar w_{ij}^{\mathrm{TE}}+(1-\beta)\bar w_{ij}^{\mathrm{qty}},\label{eq:combined_weight}\\
\mathcal L_{\mathrm{prior}}&=\frac1{|\mathcal{P}|}\sum_{(i,j)\in\mathcal{P}}w_{ij}\BCE(P_{ij},y_{ij}),\label{eq:te_loss}
\end{align}
where $\BCE(p,y)=-y\log p-(1-y)\log(1-p)$ with numerical stabilization. We use $\beta=0.5$, reserving $\theta$ for backbone parameters. Algorithm~\ref{alg:train} summarizes predictor learning.

\subsection{Prior-guided autoregressive decoding}
\label{sec:injection}

Let $\mathcal S_t\subseteq\Vc$ contain the customers selected since the most recent depot visit. Candidate compatibility is
\begin{equation}
r_t(j)=\begin{cases}
|\mathcal S_t|^{-1}\displaystyle\sum_{i\in\mathcal S_t}P_{ij},&j\in\Vc,\ |\mathcal S_t|>0,\\[3pt]
0,&j=0\ \text{or}\ |\mathcal S_t|=0.
\end{cases}
\label{eq:prior_score}
\end{equation}
The average queries consistency with the accumulated customer set, and does not depend on its enumeration. With prior strength $\lambda\ge0$, fusion gives
\begin{equation}
\widetilde u_t(j)=u_t(j)+\lambda r_t(j),\qquad
p_t(j)=\frac{\exp(\widetilde u_t(j))}{\sum_{k\in\mathcal F_t}\exp(\widetilde u_t(k))},\quad j\in\mathcal F_t.
\label{eq:logit_fusion}
\end{equation}
Infeasible actions receive zero probability. A depot action resets $\mathcal S_t$; the base solver retains its capacity, served-customer and route-closure rules. The predictor parameters are shared across instances, while $\mathbf{P}$ is computed anew for each instance. Algorithm~\ref{alg:decode} gives the decoding procedure.

At a fixed state, fusion increases relative preference for candidates with higher compatibility. Since $r_t(0)=0$, the depot logit is unchanged, but its normalized probability can decrease as customer logits increase. The odds relation and a bound on this fixed-state effect are derived in~\ref{app:fusion_bound}.

\subsection{Computational complexity}

For fixed embedding and hidden dimensions, dense pair prediction requires $O(n^2)$ evaluations and $O(n^2)$ memory for $\mathbf{P}$. Intermediate pair features can increase the practical peak memory. The prior is computed once per instance and can be shared across decoding starts.

Route aggregation can maintain $a_t(j)=\sum_{i\in\mathcal S_t}P_{ij}$. Selecting customer $i$ adds row $\mathbf{P}_{i,:}$; selecting the depot resets the accumulator. This equivalent incremental implementation costs $O(n^2)$ per trajectory with $O(n)$ additional state, as shown in~\ref{app:incremental}. The reported wall-clock overhead is evaluated separately in Section~\ref{sec:overhead}.

\section{Experiments}
\label{sec:experiments}

\subsection{Experimental setup}
\label{sec:setup}

We evaluate CVRP50 and CVRP100 generated from the standard POMO distribution~\cite{kwon2020pomo}, using 10,000 test instances for each size. Prior learning uses 1M instances with high-quality reference routes from the LEHD data-generation procedure~\cite{luo2023lehd}. We also evaluate all 10,000 instances of XML100, a structured benchmark available through CVRPLIB~\cite{xml100}, and study 500 demand-reassigned variants of one fixed geometry.

POMO is the controlled backbone. The non-augmented POMO/TRCM pair and their eightfold-augmented pair use matched decoding configurations. Non-augmented inference retains the underlying multi-start procedure. Contextual comparisons include HGS~\cite{hgs}, LKH3~\cite{lkh3}, OR-Tools~\cite{ortools}, PtrNet~\cite{vinyals2015pointer}, AM~\cite{kool2019attention}, LEHD~\cite{luo2023lehd}, ELG~\cite{gao2023elg}, ReLD~\cite{huang2025reld}, E-GAT~\cite{lei2022egat}, and GASE~\cite{wang2024gase}. These methods differ in architecture and inference procedure; the paired POMO comparisons isolate the addition of TRCM under the reported settings.

Routing metrics are average objective (Obj), the reported relative gap to the comparison's reference solution (Gap), and inference time. Prior prediction is measured by accuracy, F1, precision and recall. Because same-route labels are imbalanced, we emphasize F1 alongside accuracy. The predictor has a hidden dimension of 256 and uses 128-dimensional encoder embeddings. Experiments run on one NVIDIA RTX 4090; Table~\ref{tab:hyper} summarizes the settings.

\begin{table}[htbp]
\centering\small
\caption{Experimental settings for route-consistency learning and inference.}
\label{tab:hyper}
\begin{tabularx}{0.91\linewidth}{lX}
\toprule
Setting & Value \\
\midrule
Autoregressive backbone & POMO \\
Training instances & 1M \\
Test instances & 10,000 \\
Node embedding dimension & 128 \\
Pairwise predictor & Two-layer MLP, hidden dimension 256 \\
Optimizer & Adam \\
Learning rate & $1\times10^{-4}$ \\
Training epochs & 40 \\
Batch size & 1024 \\
TE temperature $\alpha$ & 1.0 \\
Reweighting balance $\beta$ & 0.5 \\
Prior strength $\lambda$ & 1.0 \\
Augmentation & Rotation, reflection, shrinking, noise, RA, SD \\
Inference & Non-augmented / eightfold augmentation \\
Hardware & NVIDIA RTX 4090 \\
\bottomrule
\end{tabularx}
\end{table}

\subsection{Overall performance}
\label{sec:overall}

Table~\ref{tab:cvrp_main} reports the routing comparison. Without test-time augmentation, TRCM reduces the CVRP50 gap from 3.67\% to 3.28\% and the CVRP100 gap from 2.11\% to 1.99\%. These are reductions of 0.39 and 0.12 percentage points, respectively. Average objectives decrease from 10.74 to 10.70 and from 15.84 to 15.82. The results show that the learned relation signal can complement the base policy in both non-augmented settings.

\begin{table}[htbp]
\centering\footnotesize
\setlength{\tabcolsep}{4pt}
\caption{Performance on synthetic CVRP benchmarks. Obj is average route length, Gap is the percentage gap to the comparison's reference values, and Time is inference time (s: seconds, m: minutes, h: hours). POMO/TRCM rows form the controlled backbone pairs; other methods use their respective inference procedures.}
\label{tab:cvrp_main}
\begin{tabular}{llrrrrrr}
\toprule
 & & \multicolumn{3}{c}{CVRP50} & \multicolumn{3}{c}{CVRP100} \\
\cmidrule(lr){3-5}\cmidrule(lr){6-8}
Type & Method & Obj & Gap (\%) & Time & Obj & Gap (\%) & Time \\
\midrule
Classical & HGS & 10.36 & 0.00 & 21h & 15.51 & 0.00 & 32h \\
 & LKH3 & 10.38 & 0.19 & 7h & 15.65 & 0.90 & 13h \\
 & OR-Tools & 10.98 & 5.98 & 1016s & 17.09 & 10.19 & 1054s \\
\midrule
Neural & PtrNet & 11.39 & 9.94 & $<1$s & 17.23 & 11.09 & $<1$s \\
 & AM & 10.98 & 5.98 & 3s & 16.80 & 8.32 & 8s \\
 & AM Sampling & 10.59 & 2.22 & 7m & 16.16 & 4.19 & 30m \\
 & LEHD Greedy & 10.92 & 5.41 & 2s & 16.22 & 4.58 & 30s \\
 & ELG$\times$8Aug & 10.47 & 1.06 & 10s & 15.84 & 2.13 & 3.18m \\
 & ReLD$\times$8Aug & 10.43 & 0.66 & 9s & 15.80 & 1.87 & 1.34m \\
 & E-GAT & 10.80 & 4.25 & 7s & 16.69 & 6.68 & 17s \\
 & GASE & 10.74 & 3.52 & 3s & 16.37 & 4.60 & 7s \\
\midrule
Backbone & POMO & 10.74 & 3.67 & 1s & 15.84 & 2.11 & 8s \\
 & TRCM & 10.70 & 3.28 & 3s & 15.82 & 1.99 & 15s \\
\addlinespace
Backbone$\times$8 & POMO$\times$8Aug & 10.42 & 0.58 & 26s & 15.76 & 1.61 & 5m \\
 & TRCM$\times$8Aug & 10.44 & 0.77 & 40s & 15.74 & 1.49 & 7m \\
\bottomrule
\end{tabular}
\end{table}

Eightfold augmentation produces a mixed interaction. On CVRP100, TRCM reduces the gap from 1.61\% to 1.49\% and the objective from 15.76 to 15.74. On CVRP50, the gap increases from 0.58\% to 0.77\%, and the objective increases from 10.42 to 10.44. Thus, the interaction between augmentation and route-level guidance is not uniformly beneficial. The two tested sizes also do not establish that the benefit grows with problem size.

\FloatBarrier
\subsection{Structured distribution analysis on XML100}
\label{sec:xml}

XML100 varies depot position, customer distribution, demand distribution and average route size~\cite{xml100}. We compare POMO$\times$8Aug and TRCM$\times$8Aug across all 10,000 instances. For model $m$ and group $g$, let $\mathrm{Obj}_{m,g}$ denote the reported group objective summary. The comparison is
\begin{equation}
D_g=100\,\frac{\mathrm{Obj}_{\mathrm{TRCM}\times8,g}-\mathrm{Obj}_{\mathrm{POMO}\times8,g}}{\mathrm{Obj}_{\mathrm{POMO}\times8,g}}\,\%.
\label{eq:xml_difference}
\end{equation}
Negative values favor TRCM. The factor levels are overlapping summaries of the same benchmark, rather than independent test datasets.

\begin{table}[htbp]
\centering\small
\caption{XML100 factor-wise relative objective differences for TRCM$\times$8Aug against POMO$\times$8Aug. Negative values favor TRCM. Differences are displayed to two decimal places. Each factor family summarizes the same 10,000 instances.}
\label{tab:xml100}
\begin{tabular}{llrr}
\toprule
Factor & Level & Instances & Difference (\%) \\
\midrule
Overall & All & 10000 & -0.24 \\
\midrule
Depot & Random & 3402 & -0.22 \\
 & Centered & 3322 & -0.33 \\
 & Cornered & 3276 & -0.18 \\
\midrule
Customers & Random & 3360 & -0.17 \\
 & Clustered & 3322 & -0.32 \\
 & Random-clustered & 3318 & -0.24 \\
\midrule
Demand & Unitary & 1432 & -1.04 \\
 & Small values, large CV & 1428 & -0.10 \\
 & Small values, small CV & 1428 & -0.16 \\
 & Large values, large CV & 1428 & -0.05 \\
 & Large values, small CV & 1428 & -0.23 \\
 & Quadrant-dependent & 1428 & -0.07 \\
 & Many small, few large & 1428 & -0.06 \\
\midrule
Route size & Very short & 1667 & -0.28 \\
 & Short & 1667 & -0.27 \\
 & Medium & 1667 & -0.03 \\
 & Long & 1667 & -0.12 \\
 & Very long & 1666 & -0.24 \\
 & Ultra long & 1666 & -0.53 \\
\bottomrule
\end{tabular}
\end{table}

Table~\ref{tab:xml100} shows an overall difference of $-0.24\%$, with favorable mean differences for all 19 factor levels. Across depot positions, the differences are $-0.22\%$ for random, $-0.33\%$ for centered and $-0.18\%$ for cornered depots. Clustered customers yield a larger improvement ($-0.32\%$) than random customers ($-0.17\%$), with mixed layouts at $-0.24\%$.

The largest demand-factor improvement occurs under unitary demands ($-1.04\%$). Equal demands make geometry and route organization comparatively informative for co-assignment, which is consistent with the intended structural role of the prior. Among route-size levels, ultra-long routes show the largest improvement ($-0.53\%$), while medium routes show the smallest ($-0.03\%$). A longer partial route supplies more pair relations for compatibility aggregation, but the observed trend is not monotonic. These factor summaries identify conditions in which the average benefit is larger; they do not isolate causal effects of individual factors.

\FloatBarrier
\subsection{Controlled Stability under Demand Reassignment}
\label{sec:demand_stability}

We keep customer coordinates fixed and construct 500 demand-reassigned variants of one base instance. This controlled test measures variation in solver objectives. It differs from adversarial routing or robust optimization with uncertain travel times~\cite{Zhou2024Collaboration,Xiao2024RobustNCO}.

\begin{figure}[p]
\centering
\includegraphics[width=\linewidth]{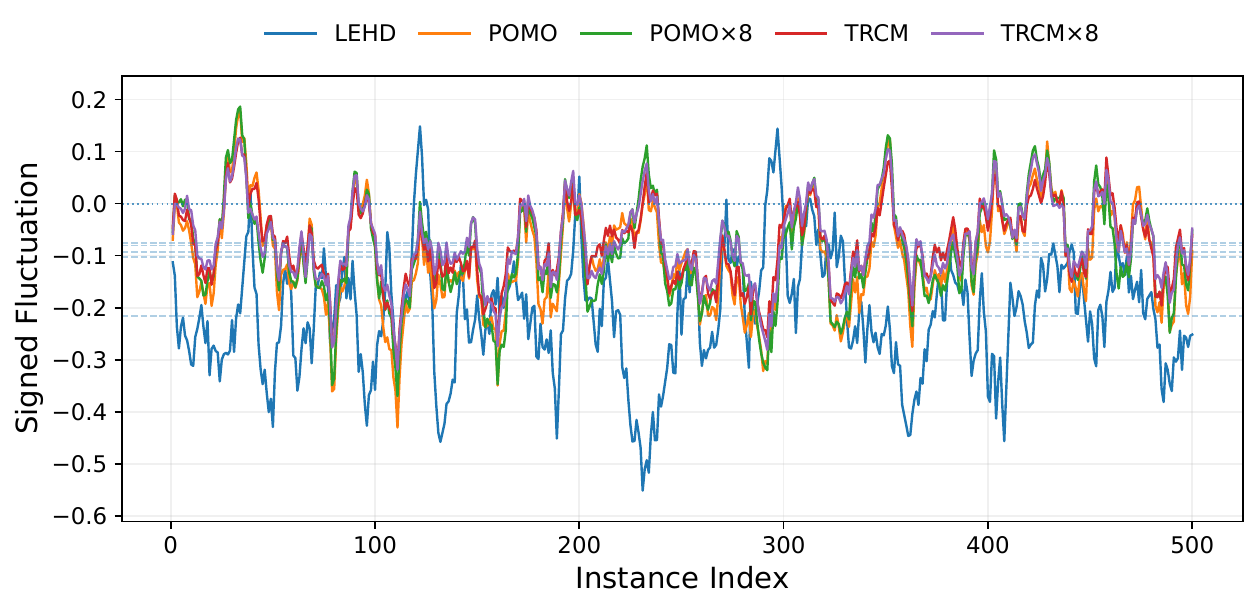}
\caption{Signed objective-fluctuation trends for 500 demand-reassigned variants of one fixed geometry. Curves show rolling means and horizontal reference lines show model-wise means.}
\label{fig:demand_signed}
\end{figure}

\begin{figure}[p]
\centering
\includegraphics[width=\linewidth]{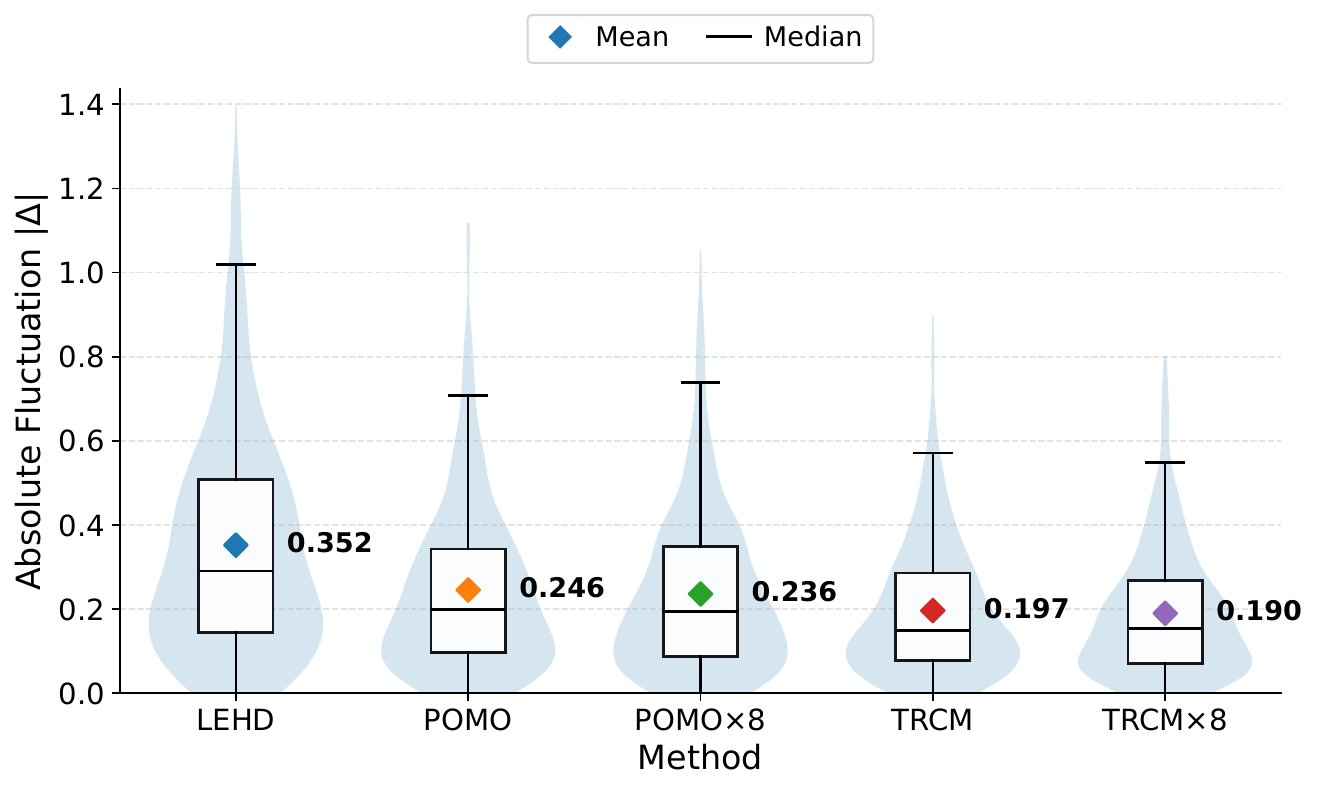}
\caption{Absolute objective-fluctuation distributions for the same variants. Diamond markers denote means and horizontal lines within boxes denote medians. TRCM has smaller mean fluctuations than the corresponding POMO settings.}
\label{fig:demand_stability}
\end{figure}

For model $m$, let $f_m^{(0)}$ be its objective on the base instance and $f_m^{(k)}$ its objective on variant $k$. Define
\begin{equation}
\Delta_m^{(k)}=f_m^{(k)}-f_m^{(0)},\qquad A_m^{(k)}=|\Delta_m^{(k)}|.
\label{eq:fluctuation}
\end{equation}
The signed fluctuation retains the direction of change, while the absolute fluctuation measures its magnitude without cancellation. Figures~\ref{fig:demand_signed} and~\ref{fig:demand_stability} display signed trends and absolute-fluctuation distributions for LEHD, POMO, POMO$\times$8, TRCM and TRCM$\times$8.

The mean absolute fluctuations are 0.352 for LEHD, 0.246 for POMO, 0.236 for POMO$\times$8, 0.197 for TRCM and 0.190 for TRCM$\times$8. TRCM therefore shows lower objective variation than the corresponding POMO settings in this controlled case. This metric measures stability relative to each solver's base objective; it does not measure optimality gaps or direct agreement between route partitions.

\FloatBarrier
\subsection{Ablation study}
\label{sec:ablation}

Table~\ref{tab:ablation_prior} evaluates prior prediction. The full configuration reaches F1 0.8065. Removing RA, SD or DA lowers F1 to 0.7841, 0.7825 or 0.7886, respectively. Removing TE produces the largest decrease, to 0.7560, supporting the contribution of topology-aware weighting in the evaluated predictor.

\begin{table}[htbp]
\centering\small
\caption{Ablation of prior-prediction training. $\Delta$F1 is the change from the full model.}
\label{tab:ablation_prior}
\begin{tabular}{lrrrrr}
\toprule
Model & Acc. & F1 & Prec. & Rec. & $\Delta$F1 \\
\midrule
TRCM (full) & 0.9336 & 0.8065 & 0.8002 & 0.8131 & 0.0000 \\
w/o RA & 0.9147 & 0.7841 & 0.6890 & 0.9096 & -0.0224 \\
w/o SD & 0.9249 & 0.7825 & 0.7720 & 0.7933 & -0.0240 \\
w/o DA & 0.9355 & 0.7886 & 0.8923 & 0.7065 & -0.0179 \\
w/o TE & 0.9111 & 0.7560 & 0.7099 & 0.8085 & -0.0505 \\
\bottomrule
\end{tabular}
\par\smallskip
\begin{minipage}{0.94\linewidth}\footnotesize
RA: route-refactor augmentation; SD: within-route demand shuffling; DA: geometric augmentation.\end{minipage}
\end{table}

The precision--recall profiles reveal complementary effects. Without RA, recall increases to 0.9096 but precision falls to 0.6890. Without DA, precision reaches 0.8923 while recall falls to 0.7065; accuracy is slightly higher than in the full configuration, despite the lower F1. The ablation thus characterizes recovery of the reference relation under imbalanced supervision. Its F1 differences concern predictor quality and are not direct estimates of downstream routing gains.

\FloatBarrier
\subsection{Hyperparameter sensitivity}
\label{sec:sensitivity}

\begin{figure}[p]
\centering
\includegraphics[width=0.73\linewidth]{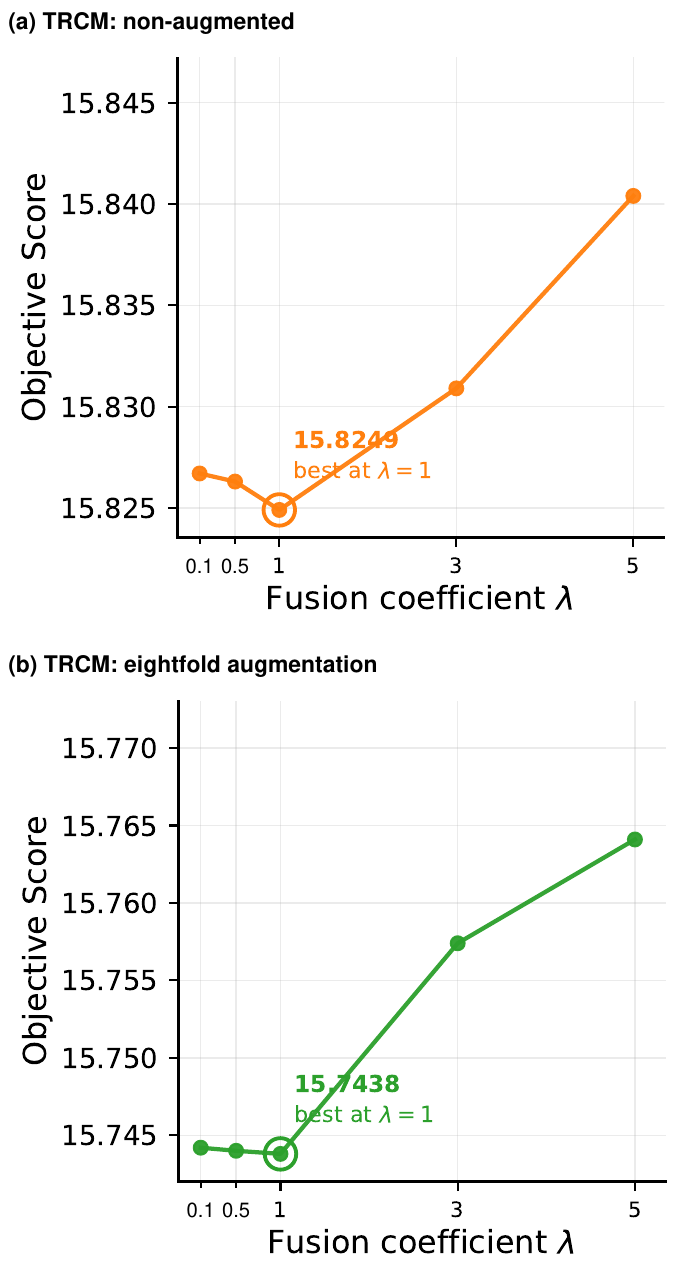}
\caption{CVRP100 objective sensitivity to fusion strength: (a) non-augmented TRCM; (b) eightfold-augmented TRCM. Both settings have their best displayed value at $\lambda=1$.}
\label{fig:lambda}
\end{figure}

Figure~\ref{fig:lambda} varies $\lambda\in\{0.1,0.5,1,3,5\}$ on CVRP100. Both inference settings have their lowest displayed objective at $\lambda=1$, with annotated values 15.8249 and 15.7438 for non-augmented and eightfold-augmented TRCM. Moderate fusion is most effective in the tested grid; large coefficients deteriorate the objective. Equation~\eqref{eq:odds_change} explains why increasing $\lambda$ strengthens the influence of compatibility on action preferences, without identifying a unique cause for the deterioration.

\begin{figure}[t]
\centering
\includegraphics[width=\linewidth]{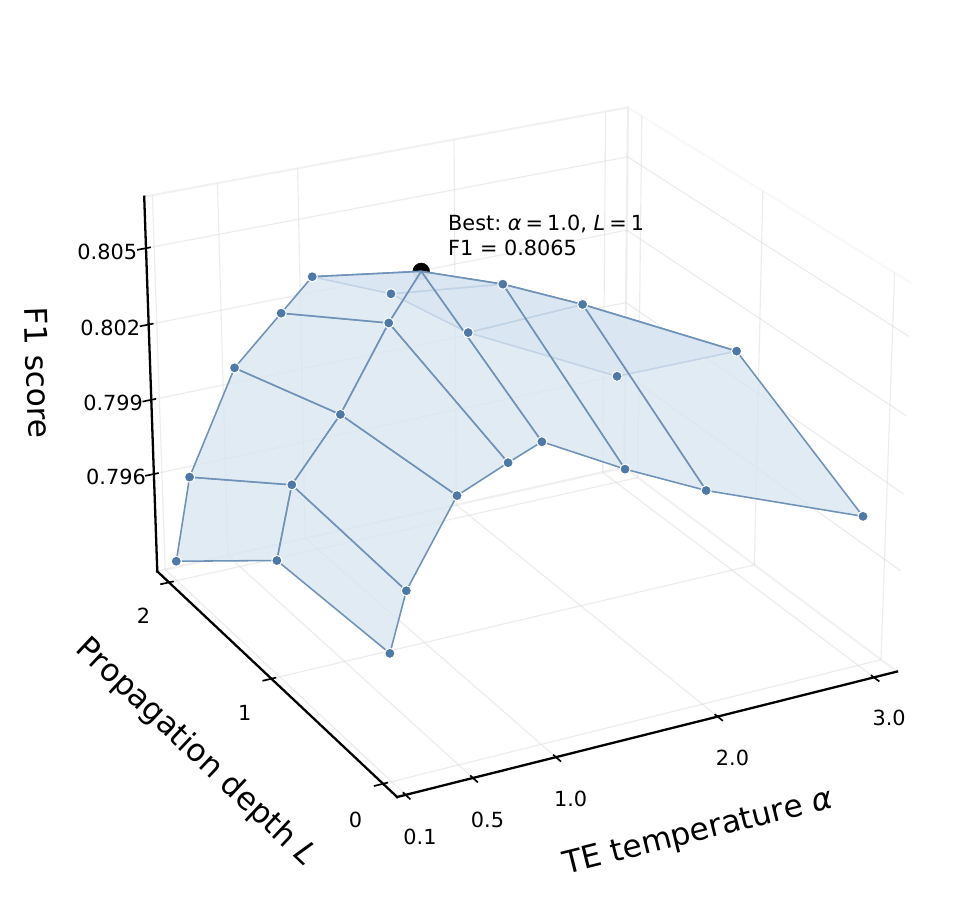}
\caption{Prior-prediction F1 as a function of TE temperature and propagation depth. The best annotated setting is $\alpha=1$, $L=1$, with F1 0.8065.}
\label{fig:te_sensitivity}
\end{figure}

Figure~\ref{fig:te_sensitivity} varies the TE temperature $\alpha\in\{0.1,0.5,1,2,3\}$ and propagation depth $L\in\{0,1,2\}$. The best annotated setting is $(\alpha,L)=(1,1)$, with F1 0.8065. Small temperatures emphasize high-entropy pairs more strongly, whereas large temperatures flatten relative weights. The observed intermediate optimum indicates a useful balance for this configuration. One propagation step improves the displayed optimum over zero steps; a second step provides no further best-score improvement. These measurements evaluate relation-prediction F1.

\FloatBarrier
\subsection{Computational overhead}
\label{sec:overhead}

Table~\ref{tab:overhead} summarizes the paired inference times. Without augmentation, time increases from 1 to 3 seconds on CVRP50 and from 8 to 15 seconds on CVRP100. With eightfold augmentation, it increases from 26 to 40 seconds and from 5 to 7 minutes, respectively. The corresponding ratios range from 1.40 to 3.00. Reference routes are prepared offline and their generation cost is excluded from inference timing.

\begin{table}[htbp]
\centering\small
\caption{Inference times under matched decoding configurations. The time ratio is TRCM time divided by POMO time.}
\label{tab:overhead}
\begin{tabular}{llrrr}
\toprule
Dataset & Setting & POMO & TRCM & Time ratio \\
\midrule
CVRP50 & Non-augmented & 1s & 3s & $3.00\times$ \\
CVRP100 & Non-augmented & 8s & 15s & $1.88\times$ \\
CVRP50 & Eightfold augmentation & 26s & 40s & $1.54\times$ \\
CVRP100 & Eightfold augmentation & 5m & 7m & $1.40\times$ \\
\bottomrule
\end{tabular}
\end{table}

These comparisons use matched decoding configurations, not matched wall-clock budgets. They quantify the cost of the prior pathway in the reported settings. Contextual times for different literature solvers are not a hardware-normalized efficiency ranking.

\section{Discussion}
\label{sec:discussion}

\paragraph{Complementary structural information}
Route membership and local transitions answer different questions. The decoder evaluates the next visit given the current state, while the prior estimates compatibility with the customers accumulated in the current route. Nonadjacent customers contribute to this compatibility. The soft-prior interface allows these signals to cooperate without requiring the predicted matrix to be decoded into a hard partition.

\paragraph{Distribution and augmentation effects}
The XML100 patterns are consistent with co-assignment being more useful when geometry or accumulated route context is informative, as in clustered, unitary-demand and ultra-long-route groups. On CVRP50, TRCM with eightfold augmentation performs worse than the corresponding POMO setting, indicating that route-level guidance and augmented inference are not uniformly complementary.

\paragraph{Reference-partition dependence}
Supervision is induced by a chosen reference solution. Different near-optimal solutions may assign some customers to different routes, so agreement with one reference partition and solution quality are related but distinct objectives. This motivates retaining a soft relation representation and evaluating the complete solver separately from its pair classifier.

\paragraph{Quality--cost trade-off}
Dense pair prediction makes the relation available for any partial-route query, at quadratic computation and storage cost. Sharing the instance prior across decoding starts and updating route scores incrementally can avoid redundant work. The measured overhead nevertheless remains material, and should be considered alongside the modest routing-quality differences.

\paragraph{Scope of controlled stability}
The demand-reassignment study shows lower objective variation for one fixed geometry; its 500 variants characterize stability within this controlled setting. Further validation across backbones, scales and independent runs can clarify the wider applicability of route-consistency guidance.

\section{Conclusion}
\label{sec:conclusion}

We presented TRCM, an explicit route-membership pathway for autoregressive neural CVRP construction. Symmetric pair prediction models customer co-assignment, topology-aware supervision addresses relation-learning imbalance, and current-route aggregation connects the learned prior to node-level decisions. The POMO-based evaluation shows improvements in both non-augmented synthetic settings and on average across XML100, a mixed interaction with eightfold augmentation, and lower objective variation in a controlled demand-reassignment case.

These results support route membership as a complementary structural signal whose value depends on the problem distribution and decoding configuration. Reference dependence and the cost of dense pair prediction remain practical considerations. Future work can examine how to improve the reliability and computational efficiency of this signal while extending its evaluation to broader routing settings.

\section*{Acknowledgements}
This work was partially supported by the National Natural Science Foundation of China under Grant No.\ 62502345, the Natural Science Foundation of Shanghai under Grant No.\ 25ZR1402511, and the Fundamental Research Funds for the Central Universities.

\section*{Data availability}
The XML100 instances and reference solutions are publicly available through CVRPLIB at \url{https://galgos.inf.puc-rio.br/cvrplib/en/xml100}. Synthetic CVRP instances and reference-route supervision follow the procedures described in Section~\ref{sec:setup}.

\appendix
\section{Analytical properties}
\label{app:analysis}

\subsection{Invariance of route membership}

\begin{proposition}
Permuting customers within each reference route or permuting the routes leaves the same-route labels in Eq.~\eqref{eq:same_route_label} unchanged. For fixed encoder representations, Eq.~\eqref{eq:pair_feature} gives $P_{ij}=P_{ji}$. For fixed $\mathbf{P}$ and a fixed current-route customer set, Eq.~\eqref{eq:prior_score} is invariant to the order in which that set is enumerated.
\end{proposition}
\begin{proof}
The label depends only on membership of $i$ and $j$ in the same customer set $\mathcal{C}(R)$. Neither permutation changes that set relation. The feature sum, absolute difference, and element-wise product are each symmetric in their two inputs, so their concatenation and deterministic predictor output are symmetric. Finally, a finite sum over $\mathcal{S}_t$ is independent of its enumeration.
\end{proof}

These properties do not imply invariance of the base logits to the entire decoding history. Two trajectories with the same customer set can still have different current nodes or decoder states, so their final fused distributions can differ.

\subsection{Relative preferences and a bound on additive fusion}
\label{app:fusion_bound}

At a fixed decoding state, let $p_t^{\mathrm{base}}$ denote the unfused distribution over the same feasible set. Softmax normalization cancels in each pairwise odds ratio, so Eq.~\eqref{eq:logit_fusion} gives
\begin{equation}
\log\frac{p_t(j)}{p_t(k)}-\log\frac{p_t^{\mathrm{base}}(j)}{p_t^{\mathrm{base}}(k)}
=\lambda\bigl(r_t(j)-r_t(k)\bigr).
\label{eq:odds_change}
\end{equation}
Thus, greater compatibility increases relative preference. The following bound controls the change in an individual feasible action's probability.

\begin{proposition}
At a fixed decoding state and feasible set, assume $r(j)\in[0,1]$ and $\lambda\ge0$. Let $p^{\mathrm{base}}(j)=\exp(u(j))/Z$ and $p(j)=\exp(u(j)+\lambda r(j))/Z'$. Then
\begin{equation}
\left|\log\frac{p(j)}{p^{\mathrm{base}}(j)}\right|\le\lambda,
\qquad
\left|\log_{10}\frac{p(j)}{p^{\mathrm{base}}(j)}\right|\le\frac{\lambda}{\ln 10}.
\end{equation}
\end{proposition}
\begin{proof}
Since $1\le\exp(\lambda r(j))\le\exp(\lambda)$, summing over the same feasible set gives $Z\le Z'\le\exp(\lambda)Z$. Therefore $0\le\log(Z'/Z)\le\lambda$. Substituting into $\log[p(j)/p^{\mathrm{base}}(j)]=\lambda r(j)-\log(Z'/Z)$ gives a value in $[-\lambda,\lambda]$. Division by $\ln 10$ gives the base-10 bound.
\end{proof}

For the depot, $r(0)=0$ and $p(0)/p^{\mathrm{base}}(0)=Z/Z'\le1$. If a feasible customer has positive score, the inequality is strict. Thus, an unchanged depot logit can accompany a lower depot probability. For a constant customer score $c$, relative customer odds are unchanged, but each customer/depot odds ratio is multiplied by $\exp(\lambda c)$. These are local algebraic statements; they do not impose a route-level performance ordering.

\subsection{Complete relation-graph limiting case}
\label{app:te_limit}

\begin{proposition}
If the TE propagation graph is complete and unweighted, with a self-loop at every node, its row-normalized adjacency is $\widetilde{\mathbf{A}}=\mathbf{1}\mathbf{1}^{\mathsf T}/n$. For any initial relation distributions, after one propagation step all node distributions coincide. Consequently, all pair TE values are equal and all normalized TE weights equal one.
\end{proposition}
\begin{proof}
Each row of $\widetilde{\mathbf{A}}$ is $\mathbf{1}^{\mathsf T}/n$, so every row of $\mathbf{Q}^{(1)}$ is the mean of the rows of $\mathbf{Q}^{(0)}$. Further propagation leaves this shared row unchanged. Averaging any two endpoints therefore produces the same vector $\mathbf{g}_{ij}^{(L)}$ for $L\ge1$, giving identical entropies and identical unnormalized weights. Dividing by their mini-batch mean yields one.
\end{proof}

In this case the quantity component can still vary by class, but the topology component cannot distinguish pairs. The supervised relation graph, rather than the density of $\mathbf{P}$, determines whether TE can distinguish local label contexts.

\subsection{Incremental route aggregation}
\label{app:incremental}

For a current route of length $s$, direct computation of every candidate score costs $O(ns)$. Summing over steps yields $O(n^3)$ in the unrestricted worst case. Alternatively, keep a vector $\mathbf{a}$ and a route-customer count $s$. When customer $i$ is selected, set $\mathbf{a}\leftarrow\mathbf{a}+\mathbf{P}_{i,:}$ and $s\leftarrow s+1$; query $r(j)=a_j/s$. On a depot action reset $\mathbf{a}=\mathbf{0}$ and $s=0$. There are $n$ customer updates and at most a linear number of nonempty-route closures, giving $O(n^2)$ aggregation work per trajectory and $O(n)$ additional state. These costs describe the equivalent incremental evaluation of Eq.~\eqref{eq:prior_score}.

\section{Learning and decoding procedures}
\label{app:algorithms}

The algorithms specify the prior-head update and its decoder interface. The supervised pair set $\mathcal{P}$ and its relation graph are training inputs; the dense prior matrix $\mathbf{P}=[P_{ij}]$ is predicted for each inference instance. Feasibility masking, action selection, state updates, and final route closure use the base solver's conventions.

\begin{algorithm}[htbp]
\caption{Topology-aware prior-head learning}
\label{alg:train}
\small
\begin{algorithmic}[1]
\Require Training instances and routes $(\mathcal{G},\mathcal{R}^{\star})$; encoder $\operatorname{Enc}_\theta$; predictor $g_\phi$; supervised pair set $\mathcal{P}$ and relation graph; $L,\alpha,\beta$
\Ensure Learned predictor $g_\phi$
\For{each training mini-batch}
\State Apply reference-label-preserving augmentation to the batch
\State $\mathbf{H}\gets\operatorname{Enc}_\theta(\widetilde{\mathcal{G}})$
\State Convert the reference routes to labels $y_{ij}$ on $\mathcal{P}$
\State Predict $P_{ij}\gets g_\phi(\mathbf{z}_{ij})$ using Eq.~\eqref{eq:pair_feature}
\State Construct $\mathbf{Q}^{(0)}$ and propagate with Eq.~\eqref{eq:te_propagation}
\State Compute $\TE_{ij}^{(L)}$ using Eq.~\eqref{eq:te}
\State Compute normalized TE and quantity weights
\State $w_{ij}\gets\beta\bar w_{ij}^{\mathrm{TE}}+(1-\beta)\bar w_{ij}^{\mathrm{qty}}$
\State Evaluate $\mathcal{L}_{\mathrm{prior}}$ using Eq.~\eqref{eq:te_loss}
\State Update $\phi$ by gradient descent
\EndFor
\State \Return $g_\phi$
\end{algorithmic}
\end{algorithm}

\begin{algorithm}[t]
\caption{Customer-prior-guided autoregressive decoding}
\label{alg:decode}
\small
\begin{algorithmic}[1]
\Require Instance $\mathcal{G}$; encoder and decoder; learned $g_\phi$; $\lambda$
\Ensure A feasible depot-to-depot routing trajectory
\State $\mathbf{H}\gets\operatorname{Enc}_\theta(\mathcal{G})$
\State Predict $\mathbf{P}=[P_{ij}]\in[0,1]^{n\times n}$ over customers
\State Initialize the base solver at the depot; $\tau\gets\emptyset$; $\mathcal{S}\gets\varnothing$
\While{unserved customers remain}
\State Obtain base logits $u_t$ and feasible set $\mathcal{F}_t$ from the solver
\For{each $j\in\mathcal{F}_t$}
\If{$j\neq0$ and $\mathcal{S}\neq\varnothing$}
\State $\displaystyle r_t(j)\gets \frac{1}{|\mathcal{S}|}\sum_{i\in\mathcal{S}} P_{ij}$
\Else
\State $r_t(j)\gets0$
\EndIf
\State $\widetilde u_t(j)\gets u_t(j)+\lambda r_t(j)$
\EndFor
\State Select $a_t$ using the base solver's selection rule on masked $\softmax(\widetilde u_t)$
\State Append $a_t$ to $\tau$ and update the base solver state
\If{$a_t=0$}
\State $\mathcal{S}\gets\varnothing$
\Else
\State $\mathcal{S}\gets\mathcal{S}\cup\{a_t\}$
\EndIf
\EndWhile
\State Finalize the last depot return using the base solver's route-closure convention
\State \Return $\tau$
\end{algorithmic}
\end{algorithm}

\FloatBarrier
\bibliographystyle{elsarticle-num}
\bibliography{refs}
\end{document}